\documentclass[sigconf]{aamas} 

\usepackage{balance} 
\usepackage{times}
\usepackage{graphicx}

\usepackage{amsthm}
\usepackage{siunitx}
\usepackage{xspace}
\usepackage{caption}
\usepackage{subcaption}
\usepackage{calrsfs}
\usepackage[fleqn]{mathtools}
\usepackage{color}
\usepackage{booktabs}
\usepackage{setspace}
\usepackage{float}
\usepackage[htt]{hyphenat}
\usepackage[autostyle, english = american]{csquotes}
\usepackage{url}
\setcopyright{ifaamas}
\acmConference[AAMAS '23]{Proc.\@ of the 22nd International Conference
on Autonomous Agents and Multiagent Systems (AAMAS 2023)}{May 29 -- June 2, 2023}
{London, United Kingdom}{A.~Ricci, W.~Yeoh, N.~Agmon, B.~An (eds.)}
\copyrightyear{2023}
\acmYear{2023}
\acmDOI{}
\acmPrice{}
\acmISBN{}

\acmSubmissionID{fp0853}

\title[POMCP+ILP]{Learning Logic Specifications for Soft Policy Guidance\\ in POMCP
}

\author{Giulio Mazzi}
\affiliation{
  \institution{University of Verona}
  \city{Verona}
  \country{Italy}}
\email{giulio.mazzi@univr.it}
\authornote{Authors contributed equally.}
\author{Daniele Meli}
\affiliation{
  \institution{University of Verona}
  \city{Verona}
  \country{Italy}}
\email{daniele.meli@univr.it}
\authornotemark[1]
\author{Alberto Castellini}
\affiliation{
  \institution{University of Verona}
  \city{Verona}
  \country{Italy}}
\email{alberto.castellini@univr.it}
\author{Alessandro Farinelli}
\affiliation{
  \institution{University of Verona}
  \city{Verona}
  \country{Italy}}
\email{alessandro.farinelli@univr.it}

\begin{abstract}

Partially Observable Monte Carlo Planning (POMCP) is an efficient solver for Partially Observable Markov Decision Processes (POMDPs). It allows scaling to large state spaces by computing an approximation of the optimal policy locally and online, using a Monte Carlo Tree Search based strategy. 
However, POMCP suffers from sparse reward function, namely, rewards achieved only when the final goal is reached, particularly in environments with large state spaces and long horizons.
Recently, logic specifications have been integrated into POMCP to guide exploration and to satisfy safety requirements. However, such policy-related rules require manual definition by domain experts, especially in real-world scenarios. In this paper, we use inductive logic programming to learn logic specifications from traces of POMCP executions, i.e., sets of belief-action pairs generated by the planner. 
Specifically, we learn rules expressed in the paradigm of answer set programming.
We then integrate them inside POMCP to provide soft policy bias toward promising actions. In the context of two benchmark scenarios, \emph{rocksample} and \emph{battery}, we show that the integration of learned rules from small task instances can improve performance with fewer Monte Carlo simulations and in larger task instances.
We make our modified version of POMCP publicly available at \url{https://github.com/GiuMaz/pomcp_clingo.git}.
\end{abstract}


\newcommand{\BibTeX}{\rm B\kern-.05em{\sc i\kern-.025em b}\kern-.08em\TeX}
\newcommand{\stt}[1]{{\small\texttt{#1}}}
\newcommand{\no}[1]{{\; {not} \;}}
\newcommand{\myif}[1]{{\texttt{:-}}}

\DeclareMathAlphabet{\pazocal}{OMS}{zplm}{m}{n}
\theoremstyle{definition}
\newtheorem{defn}{Definition} 

\newenvironment{frcseries}{\fontfamily{frc}\selectfont}{}

\begin{document}


\pagestyle{fancy}
\fancyhead{}


\maketitle 


\section{Introduction}
Partially Observable Markov Decision Processes (POMDPs) are a popular framework for modeling systems with state uncertainty \cite{cassandra2013incremental}. 
The state cannot be completely observed by the agent, therefore it is modeled as a probability distribution, called \emph{belief}. 
However, computing optimal POMDP policies (exploring the full belief space) is hard \cite{papadimitriou1987complexity}. 
Several approximate algorithms have been proposed to mitigate the computational issue, such as, Partially Observable Monte Carlo Planning (POMCP) \cite{silver2010monte} and Determinized Sparse Partially Observable Tree (DESPOT) \cite{somani2013despot}. 
Most existing approaches rely on exploring only a part of the belief space, e.g., with random particle sampling in POMCP; hence, a trade-off must be found between quality of policies and exploration cost. This hinders the scalability to real-world scenarios with large belief space. 
Furthermore, POMCP policies are black-box, hence the decision process underlying policy generation is not transparent to other agents and humans, reducing safety and trustability.

In this paper we address the aforementioned issues, introducing soft logic-based policy bias in the Monte Carlo tree exploration in POMCP.
Differently from existing approaches which set bounds to the exploration process \cite{mazzi2021rule, leonetti2016synthesis}, we use logic specifications to only \emph{suggest} actions to POMCP solver, based on current belief. Furthermore, we do not require specifications to be provided by a domain expert. 
Instead, we propose to learn logic specifications offline from traces of executions, i.e., sequences of belief-action pairs (generated using POMCP or other solvers).
We express specifications in Answer Set Programming (ASP) \cite{lifschitz1999answer}, a state-of-the-art paradigm for logic planning \cite{meli2023logic, erdem2018applications, ginesi2020autonomous, tagliabue2022deliberation}.
We convert the belief to ASP representation, in terms of higher-level domain \emph{features} specified by an expert user. Then, we use them to represent logic relationships between beliefs and actions.
Defining features requires only basic domain knowledge (e.g., relevant domain quantities used to represent POMDP task instances).
We then exploit Inductive Logic Programming (ILP) \cite{muggleton1991inductive} to infer ASP rules from feature-action pairs.

We validate our approach on two benchmark POMDP problems, \emph{rocksample} and \emph{battery}. 
The contributions of this paper are the following: i) we learn specifications about POMDP policies in the ASP semantics from execution traces, requiring only definition of domain-relevant features from experts; ii) we use ASP statements to guide POMCP exploration and increase performance and scalability in challenging domain instances, e.g., with increased planning horizons; iii) we evaluate the impact of the quality and amount of execution traces on learning outcome and POMCP performance. 
The implementation of our extended POMCP algorithm is made available.


\section{Related works}\label{sec:sota}
Merging MDP / POMDP solving algorithms with logic reasoning is a recent research trend \cite{leonetti2016synthesis, sridharan2019reba, mazzi2021rule}. One advantage of this approach is to guide policy search with commonsense reasoning and expert knowledge. For instance, in the REBA framework \cite{sridharan2019reba} ASP is used to describe spatial relations between objects and rooms in a houseware domain, hence driving a robotic agent to choose a specific room to inspect, while solving simpler MDP problems locally. Similarly, authors of \cite{leonetti2016synthesis} propose DARLING, which uses ASP statements to bound MDP exploration in a simulated grid domain (similar to rocksample) and in a real service robotics scenario. 

Logic statements can help also avoid unwanted behaviors, e.g. in safety-critic scenarios. To this aim, authors of \cite{mazziAAMAS2021,mazzi2021rule} refine rule templates to identify unexpected decisions and shield undesired branches of belief-action tree from exploration in POMCP, in the context of velocity regulation for a mobile robot. In \cite{wang2021online}, a goal-constrained belief space containing only safely reachable states from the initial one is defined with propositional logics, and a Satisfiability Modulo Theory (SMT) solver is used to guarantee proper execution of an houseware automation task. 

Aforementioned approaches, as well as similar ones implementing more complex reasoning with, e.g., temporal logic \cite{de2019foundations, de2020imitation, leonetti2012automatic}, have one major drawback: logic statements are assumed to be provided by the user. This is unrealistic, especially in complex domains, where even expert users can hardly define accurate policy rules.
Moreover, previous works use logic statements to model an additional reward in MDPs. 

In contrast, we learn logic relations between domain-relevant user-defined features, which are more easily available, and actions with the paradigm of ILP under ASP semantics, and we use them to directly advise POMCP simulations. POMCP was recently used successfully in real-world applications, such as robot planning \cite{pomp2020bmvc,Giuliari2021,zuccotto2022}. Learning meaningful task representations has been proposed, e.g., in \cite{bonet2020learning,rodriguez2021learning}, though computational efficiency is mined by the use of an ASP meta-program. 
ILP has been shown to be effective in enhancing comprehensibility of black-box models \cite{d2020towards, rabold2018explaining}. Recent applications in robot planning include task knowledge learning from human labelled examples with ILASP \cite{meli2021inductive}, and learning of temporal specifications from expert-generated MDP traces \cite{de2020imitation}. ILASP is more efficient than, e.g., \cite{rodriguez2021learning} since it does not rely on ASP meta-program representation as of latest versions.

\section{Problem definition and case studies}\label{sec:domains}
The goal of this paper is to discover logical rules which underlie the decision-making process of a POMDP agent, analyzing patterns of execution. In particular, the agent performs actions depending on the current state of the environment and the goal of the assigned task. Hence, we want to find logical rules which match POMDP representation of the environment to a specific policy, with the aim of advising POMCP solver and directing the execution of the agent in future online executions. 
In particular, by integrating rules in POMCP exploration we aim to improve the final cumulative reward in typically challenging task instances, e.g., with long planning horizons and when the number of available actions (and corresponding needed simulations) increases.     
We now provide an informal description of the two domains chosen as case studies, \emph{battery} and \emph{rocksample}.
In the following sections, we will refer to rocksample as a running example to explain the main elements of the background and methodology.

\subsection{Battery}
\begin{figure}
    \centering
    \begin{subfigure}{0.3\textwidth}
    \centering
    \includegraphics[width=0.9\linewidth]{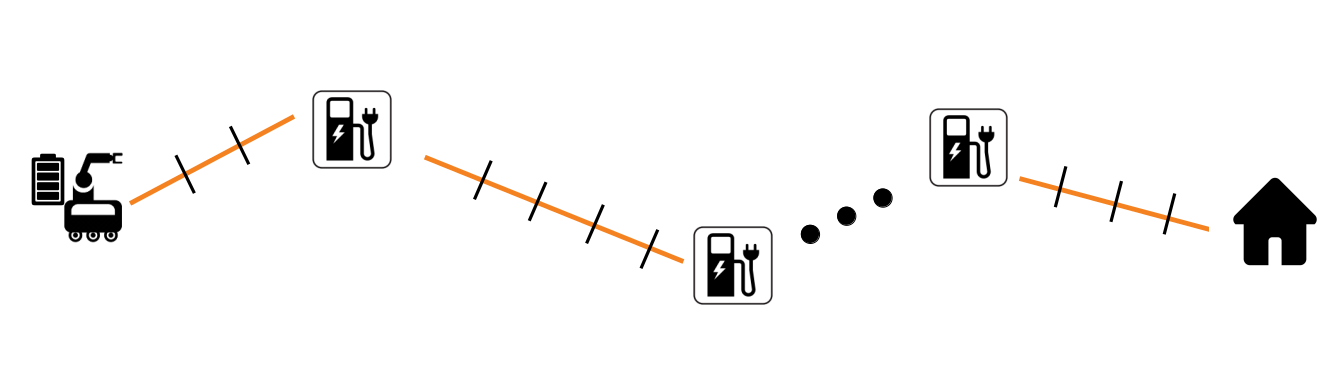}
    \caption{Battery\label{fig:setup_battery}}
    \end{subfigure}
    \begin{subfigure}{0.3\textwidth}
    \centering
    \includegraphics[width=0.6\linewidth]{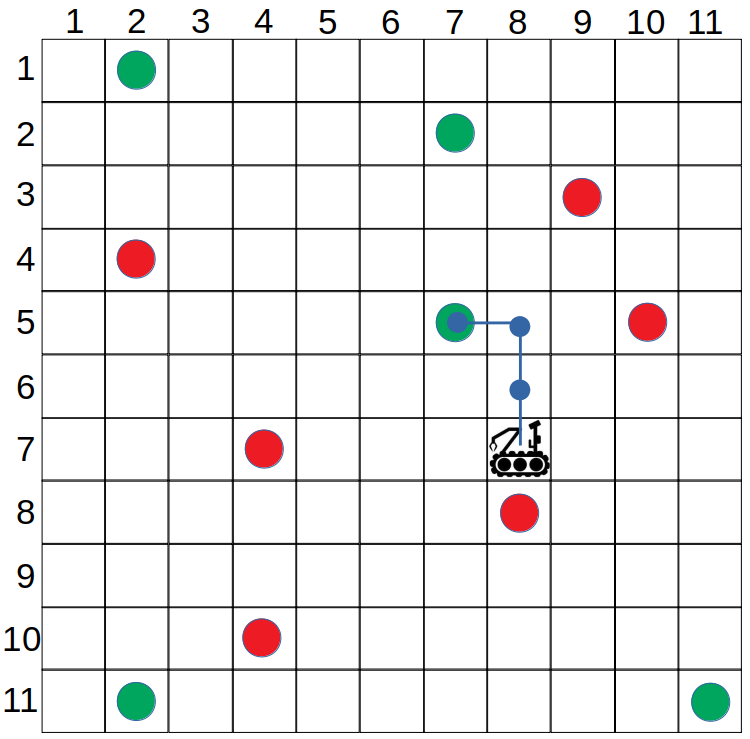}
    \caption{Rocksample\label{fig:setup_rocksmaple}}
    \end{subfigure}
    \caption{Example scenarios for our two case studies.}
    \label{fig:setup_rocksample}
\end{figure}
In the battery domain, an agent moves forward towards a target location, gaining a positive reward in case of success and a negative reward in case of battery depletion. The agent has limited energy autonomy, i.e., a battery with $L$ levels of charge. Moving may reduce the level of the battery, but at known locations on the path there are recharge stations. However, recharging has a cost (i.e., a negative reward). The observable state of the system is represented by positions of the target, agent and stations, while the unobservable part of the state is the level of the battery, which can be checked with a noisy sensor. The belief in this case is a probability distribution over the possible battery levels.
An explanatory scenario is depicted in Figure \ref{fig:setup_battery}, with segments connecting the robot, stations and the goal subdivided in discrete steps of advancement.

\subsection{Rocksample}
In the rocksample domain, an agent can move in cardinal directions (north, south, east and west) on a $NX N$ grid, one cell at a time, with the goal to reach and sample a set of $M$ rocks with known position on the grid. Rocks may be valuable or not. Sampling a valuable rock yields a positive immediate reward, while sampling a worthless rock yields negative reward. The observable state of the system is described by the current position of the agent and rocks, while the value of rocks is uncertain before sampling. The belief is in this case a probability distribution over all possible configurations of rock values. However, the agent can check the value of a rock, with an accuracy depending on the distance between the rock and the agent. Finally, the agent can obtain a positive reward exiting the grid from the right-hand side.
An explanatory scenario with $N=11$ and $M=11$ is depicted in Figure \ref{fig:setup_rocksample}b, where valuable and invaluable rocks are marked as green and red dots, respectively, and the agent is at location $(8,7)$ initially.

\section{Background}\label{sec:background}
We first introduce the definitions of POMDPs and POMCP, then we describe key concepts of  ASP and ILP which are used in the following to describe the proposed methodology.  

\subsection{Partially Observable Monte Carlo Planning}
A Partially Observable Markov Decision Process (POMDP)~\cite{Kaelbling98} is a tuple $(S, A, O, T, Z, R, \gamma)$,
where $S$ is a set of partially observable \emph{states},
$A$ is a set of \emph{actions},
$Z$ is a finite set of \emph{observations},
$T$:~$SX A \rightarrow \Pi(S)$ is the \textit{state-transition model}, with $\pazocal{B} = \Pi(S)$ probability distribution over states,
$O$:~$SX A \rightarrow \Pi(Z)$ is the \textit{observation model},
$R$ is the \textit{reward function} and $\gamma \in [0,1]$ is a \textit{discount factor}.
An agent must maximize the \emph{discounted return} $E[\sum_{t=0}^{\infty} \gamma^t R(s_t,a_t)]$.
A probability distribution over states, called \emph{belief}, is used to represent the partial observability of the true state.
To solve a POMDP it is required to find a \emph{policy}, namely a function $\pi$:~$\pazocal{B} \rightarrow A$ that maps beliefs $\pazocal{B}$ into actions.

In this work, we focus on \emph{Partially Observable Monte-Carlo Planning (POMCP)}~\cite{silver2010monte} to solve POMDPs.
POMCP is an \emph{online} algorithm that solves POMDPs by using Monte Carlo techniques.
The strength of POMCP is that it does not require an explicit definition of the transition model, observation model, and reward.
Instead, it employs a black-box simulator of the environment.
POMCP uses \emph{Monte Carlo Tree Search} (MCTS) at each time step to explore the belief space and select the best action.
\emph{Upper Confidence Bound for Trees (UCT)}~\cite{Kocsis2006} is used as a search strategy to select the subtrees to explore. In particular, given history $h$ of former belief-action pairs, UCT suggests to explore action $a$ which maximizes the action value:
\begin{equation} \label{eq:UCT}
V_{UCT}(ha) = V(ha) + c \cdot \sqrt{\frac{\log N(h)}{N(ha)}}
\end{equation}
where $V(ha)$ is the expected return achieved by selecting action $a$, $N(h)$ is the number of simulations performed from history $h$, $N(ha)$ is the number of simulations performed while selecting action $a$, and $c$ is known as the exploration constant.
UCT is used to balance the exploration of new actions (i.e., action with low $N(ha)$) and the exploitation of effective actions (i.e., action with high $V(ha)$).
The belief is implemented as a \emph{particle filter}, which is a sampling over the possible states. The belief can be initialized randomly or considering prior knowledge about the environment \cite{Castellini2019}. 
At each time step, a particle (representing a specific state of the POMDP) is selected from the filter and used as an initial point to perform a simulation in Monte Carlo tree.
The particle filter is then updated at each step after an action is performed.
A simulation is a sequence of action-state pairs that collects a discounted return. Running different simulations, we can estimate and select the action which leads to the highest return.
If required, new particles can be generated from the current state through a process of \emph{particle reinvigoration}.

\subsection{Answer Set Programming}\label{sec:asp_theory}
An ASP program represents a domain of interest with a \emph{signature} and \emph{axioms} \cite{calimeri2020asp}. The signature is the alphabet of the domain, defining its relevant attributes as variables (with discrete ranges) and predicates of variables (\emph{atoms}).
For example, in the rocksample domain variables of interest are: rock identifiers \stt{R} (natural), distances between agent and rock positions \stt{D} (integer), and (discretized) probabilities computed from the belief \stt{V}$\in \{0, 10, ..., 100\}$\footnote{From now on, probability values represent percentages, e.g., $\{0\%, 10\%, ..., 100\%\}$.}.
Atoms typically represent environmental features and actions.
Possible atoms in rocksample are \stt{guess(R,V)} and \stt{sample(R)}, denoting respectively, the probability \stt{V} the rock \stt{R} is valuable, and the action of sampling rock \stt{R}.
A variable whose value is assigned is said to be \emph{ground} (e.g., \stt{R}=1). An atom is ground if its variables are ground. Ground atoms become Booleans with truth value.

Axioms are logical relations between atoms. 
In this paper, we consider only causal rules and weak constraints.
A \emph{causal rule} \stt{h :- b$_{1..n}$} defines preconditions as the logical conjunction $\bigwedge_{i=1}^n$\stt{b$_i$} (\emph{body} of the rule) for the \emph{head} \stt{h}.
For instance,
\begin{equation}
    \label{eq:asp_ex}\stt{sample(R) :- guess(R,V), V>60.}
\end{equation}
\noindent
means that rock \stt{R} can be sampled if the agent believes it is valuable with probability \stt{V}$>60\%$.
\emph{Weak constraints} express preference criteria between atoms.
A weak constraint has the form:
\begin{equation*}
    :\sim \stt{b}_1\stt{(V}_1, \ldots, \stt{V}_n\stt{)}, \ldots, \stt{b}_m\stt{(V}_1, \ldots, \stt{V}_n\stt{)}.\stt{[w@l, V}_1, \ldots , \stt{V}_n\stt{]}
\end{equation*}
where \stt{w} is the weight, \stt{l} the (integer) priority level (used if multiple weak constraints are specified), \stt{b}$_i$ are atoms and \stt{V}$_j$ variables. 
The weight can either be one variable among \stt{V}$_j$'s, or an integer. If the weight is integer, the weak constraint means that grounding of atoms \stt{b}$_i$'s should be penalized. If the weight is a variable, say \stt{V$_1$}, then the constraint means that lower values for \stt{V$_1$} should be preferred.
As an example, if \eqref{eq:asp_ex} holds and the following weak constraint is specified:
\begin{equation}
    \label{eq:asp_th_wc}:\sim \stt{sample(R), guess(R,V). [-V@1, R, V]}
\end{equation}
\noindent
when $\geq 2$ rocks have probability $>60\%$ to be valuable, the one with higher $V$ (negative weight) will be preferred.

Given an ASP task description, an ASP solver computes \emph{answer sets}. An answer set is the minimal set of ground atoms satisfying axioms. 
Starting from an initial grounding, the solver deduces ground heads of rules from known ground body atoms. 
For the scope of this paper, answer sets contain ground features and actions.
For instance, with reference to \eqref{eq:asp_ex}, if at a specific step the agent's belief distribution (formalized in ASP) is \{\stt{guess(1,50), guess(2,70)}\} (namely, the agent believes rock 1 is valuable with probability 50\%, and rock 2 is valuable with probability 70\%), then action \stt{sample(2)} will be grounded, and the answer set will contain all three ground atoms, i.e., \{\stt{guess(1,50), guess(2,70), sample(2)}\}.

\subsection{Inductive Logic Programming}\label{sec:ilasp_theory}
An ILP problem $\pazocal{T}$ under ASP semantics is defined as the tuple $\pazocal{T} = \langle B, S_M, E \rangle$, where $B$ is the \emph{background knowledge}, i.e. a set of atoms and axioms in ASP syntax (e.g., ranges of variables); $S_M$ is the \emph{search space}, i.e. the set containing all possible ASP axioms that can be learned; and $E$ is a set of \emph{examples} (e.g., a set of ground atoms constructed from traces of execution). 
Our goal is to construct a set of ASP axioms belonging to $S_M$, called \emph{hypothesis} $H$, that can explain as many as possible of the examples in $E$. To this end, we use state-of-the-art ILASP learner \cite{ILASP_system}, where examples are \emph{Context-Dependent Partial Interpretations} (CDPIs).
\begin{defn}[Partial interpretation]
\label{def:partial_int}
Let $P$ be an ASP program. A \emph{partial interpretation} of $P$ is defined as $e = \langle e^{inc}, e^{exc} \rangle$, where $e^{inc}$ is named \emph{included set}, i.e., a subset of ground atoms which can be part of an answer set of $P$; $e^{exc}$ is named \emph{excluded set}, i.e., a subset of ground atoms which are not part of an answer set of $P$.
\end{defn}
\begin{defn}[Context-dependent partial interpretation (CDPI)]
\label{def:CDPI}
A CDPI of an ASP program $P$ is a tuple $\langle e, C \rangle$, where $e$ is a partial interpretation of $P$ and $C$ is a set of atoms called \emph{context}.
\end{defn}
\noindent
In this paper, partial interpretations contain atoms for actions, while the context involves feature atoms. 
In this way, policy specifications can be learnt.
In our formulation, $e^{inc}$ includes observed actions and $e^{exc}$ includes unobserved (hence not executed) actions.
For instance, in rocksample, a CDPI may be
\begin{equation*}
    \langle \langle \{\stt{sample(2)}\}, \{\stt{sample(1)}\} \rangle, \{\stt{guess(1,50), guess(2,70)}\} \rangle
\end{equation*}
\noindent
where $C=$\{\stt{guess(1,50),}
\stt{guess(2,70)}\}, $e^{exc}=$\{\stt{sample(1)}\}, $e^{inc}=$\{\stt{sample(2)}\}.
The meaning of this CDPI is that, given the context defining the current belief of the agent (namely, it believes that rock \stt{1} is valuable with probability 50\% and rock \stt{1} is valuable with probability 70\%), rock \stt{2} should be sampled and rock \stt{1} should not.

We can now define the ILASP problem considered in this paper, hence the properties to be satisfied by $H$:
\begin{defn}[ILASP task with CDPIs]
\label{def:ILASP_CDPI}
An ILASP learning task with CDPIs is a tuple $\pazocal{T} = \langle B, S_M, E \rangle$, where $E$ is a set of CDPIs such that:
\begin{equation*}
     \forall e = \langle \langle e^{inc}, e^{exc} \rangle, C \rangle \in E : \ B \cup H \cup C \models e^{inc} \land \ B \cup H \cup C \not\models e^{exc}
\end{equation*}
\end{defn}
\noindent
In other words, ILASP finds axioms which guarantee that actions in $e^{inc}$, observed in the examples, can be executed (i.e., can be grounded in an answer set), while unobserved actions in $e^{exc}$ cannot, given the context set of environmental features.
In addition, ILASP finds the \emph{minimal} hypothesis $H$, i.e., axioms with the least number of atoms. 
This increases the comprehensibility of learned axioms and improves efficiency of ASP solving. ILASP can also learn weak constraints from \emph{ordered CDPIs}, i.e., partial interpretations with pre-defined preference values \cite{LRB16}.

Finally, ILASP finds the hypothesis which explains \emph{most} of the examples, i.e., there may be CDPIs which are not covered by $H$. When a hypothesis is found, ILASP also returns the number of non-convered CDPIs, which quantifies the reliability of $H$ with respect to the provided set $E$.

\section{Methodology}\label{sec:methods}
We now describe our methodology for learning policy-related logic specifications from POMDP traces of execution, for a given domain of interest.
In particular, we want to represent the policy map $\pi : \pazocal{B} \rightarrow A$ as a set of logical formulas, through a new map $\Gamma : \pazocal{F} \rightarrow \pazocal{A}$, being \stt{$\pazocal{F} = \{$F$_i\}$} a set of categorical environmental features defined by an user (e.g., \stt{guess(R,V)} in rocksample), and \stt{$\pazocal{A} = \{$A$_i\}$} the logical formulation of $A$ (containing, e.g., \stt{sample(R)} for rocksample). 
Assuming actions and features are represented as ASP atoms, three main steps are required to build map $\Gamma$ and integrate logical formulas into POMCP: \emph{ASP representation of the domain} (Section \ref{sec:asp_methods}); \emph{definition of the ILASP problem} starting from traces of POMDP executions (Section \ref{sec:ilasp_methods}); \emph{integration of learned axioms in POMCP} (Section \ref{sec:pomcp_methods}).

\subsection{ASP representation of the domain}\label{sec:asp_methods}
In order to represent features and actions in ASP syntax, we define a \emph{feature map} $F_\pazocal{F} : \pazocal{B} \rightarrow G(\pazocal{F})$ and an \emph{action map} $F_\pazocal{A} : A \rightarrow G(\pazocal{A})$, being $G(\cdot)$ the \emph{grounding function} defining all possible groundings of an atom or a set of atoms. 
For instance, considering $\pazocal{A}=$ \{\stt{sample(R)}\} in a rocksample scenario with 2 rocks, $G(\pazocal{A})=$ \{\stt{sample(1), sample(2)}\}.
Once maps $F_\pazocal{F}, F_\pazocal{A}$ are defined by an user, it is possible to automatically translate traces of POMDP executions (i.e., sequences of belief-action pairs) to sets of ground atoms.

\subsection{ILASP problem definition from POMDP traces}\label{sec:ilasp_methods}
We can now define an ILASP task as in Section \ref{sec:ilasp_theory}, with examples built from POMDP traces using maps $F_\pazocal{F}, F_\pazocal{A}$ and target hypothesis $H = \Gamma$.
We make the following assumptions to define the ILASP learning problem:
\begin{itemize}
    \item examples are taken from ``good'' traces of execution, i.e., where the policy is nearly optimal, so the agent successfully completed the task with high discounted return. In this way, learned rules are more efficient in online execution. In Section \ref{sec:exp} we detail how POMDP traces for learning are generated in order to satisfy this assumption (e.g., tuning POMCP parameters);
    \item for each action, the corresponding map $\Gamma$ does not depend on maps for other actions, i.e., axioms for each action are independent on axioms for other actions (e.g., rules for checking or sampling a rock are mutually independent, but are only connected to features). Hence, we can define a separate ILASP task for each action, increasing computational efficiency thanks to the shrinkage of the search space.
\end{itemize}
Each ILASP problem is defined by background knowledge $B$, search space $S_M$ and examples $E$.
In this paper, $B$ only contains the definitions of ASP variables and ranges.
Since $\Gamma : \pazocal{F} \rightarrow \pazocal{A}$, we define $S_M$ allowing only atoms $\in \pazocal{A}$ in the head of candidate axioms and atoms $\in \pazocal{F}$ in the body. 
Examples are extracted from traces of POMDP executions using maps $F_{\pazocal{F}}$ and $F_{\pazocal{A}}$.
Specifically, for each belief-action $\langle b, a \rangle \in \pazocal{B} X A$, we obtain a pair of ground feature set and action $\langle \stt{belief}, \stt{action} \rangle$, where \stt{belief}$\subseteq G(\pazocal{F})$ and \stt{action}\stt{$\subseteq G($A$_i)$}, \stt{A$_i$} being the ASP atom representing $a$. 
For instance, in a rocksample instance with 2 rocks, assume sampling of the second rock is performed, and the probabilities that rocks 1 and 2 are valuable are 70\% and 80\%, respectively.
Then, \stt{A$_i=$sample(R)}, \stt{action}$=$\stt{sample(2)} and \stt{belief}$=$\{\stt{guess(1,70), guess(2,80)}\}.
We then generate a CDPI in the form
\begin{align}
    \label{eq:pos}\langle \langle \{\stt{action}\}, G(\stt{A}_i)\setminus \{\stt{action}\} \rangle, \stt{belief} \rangle
\end{align}
\noindent
where \stt{$G($A$_i) \setminus\{$action$\}=\{$sample(1)$\}$} in the previous example.
Moreover, at each step where an action $a$ is not executed, we define an additional CDPI
\begin{equation}
    \label{eq:cex}\langle \langle \emptyset, G(\stt{A}_i) \rangle, \stt{belief} \rangle
\end{equation}
For instance, in the previous example, if $a$ is the checking action, $G(\stt{A}_i) = \{\stt{check(1), check(2)}\}$, and similarly for other actions.
In this way, learnt axioms will provide more meaningful and useful policy specifications, since they will be induced also from \emph{counterexamples} (i.e., examples where actions are not executed).

In order to learn weak constraints (i.e., preferences on specific actions with respect to others), for each example as \eqref{eq:pos} we generate CDPIs in the form:
\begin{equation}
    \label{eq:pref}\langle \langle G(\stt{A}_i)\setminus \{\stt{action}\}, \emptyset \rangle, \stt{belief} \rangle
\end{equation}
\noindent
In other words, all non-executed actions are in $e^{inc}$, while $e^{exc} = \emptyset$.
We then specify a preference of \eqref{eq:pos} over \eqref{eq:pref}. 
This is needed, since execution traces only contain a single action at each step, but possibly others could be executed.
For instance, consider the former rocksample instance and assume \eqref{eq:asp_ex} holds. Sampling of both rocks is possible, since they are valuable with probability $\geq 60\%$, but only the second rock was actually sampled. 
We then define the following examples, corresponding respectively to \eqref{eq:pos}-\eqref{eq:pref}:
\begin{subequations}
\begin{align}
    \label{eq:pref1}&\langle \langle \{\stt{sample(2)}\}, \{\stt{sample(1)}\} \rangle, \{\stt{guess(1,70), guess(2,80)}\} \rangle\\
    \label{eq:pref2}&\langle \langle \{\stt{sample(1)}\}, \emptyset \rangle, \{\stt{guess(1,70), guess(2,80)}\} \rangle
\end{align}
\end{subequations}
Then, we specify that \eqref{eq:pref1} must be preferred to \eqref{eq:pref2}.

For each action, we consider hypothesis which covers most of the examples and use it to bias POMCP exploration, as explained in next subsection.


\subsection{Integrating axioms in POMCP}\label{sec:pomcp_methods}
We integrate the computed axioms inside POMCP to help the algorithm selecting the best action given the current state and belief.
Specifically, when a new node is reached during the Monte Carlo Tree Search (MCTS), we instantiate the variables of the axioms with values computed from the current belief.
POMCP represents the belief of the root node using a particle filter (i.e., the algorithm keeps a finite collection of possible states for each belief). Thus, we compute the values required to ground ASP features by analyzing all the particles. For example, to compute \stt{guess(R,V)} for rock \stt{1} we compute the percentage of particles in which rock \stt{1} is valuable. 
When we reach a new non-root node, no particle filter is available. However, we still ground features that depend only on the observable part of the state (e.g., the current position in rocksample).
These features need to be re-grounded only for previously visited non-root nodes that become a root (i.e., we move the simulation one step forward by selecting an action and receiving an observation, and the new root is a node that we visited previously during a simulation). In this case, the features are re-instantiated using the new full belief. In all the other cases, no recomputation is needed since the observable part of a node never changes.
When all features of a node are instantiated, we call the ASP solver to ground new actions.

To combine logical rules with POMCP, we introduce a prior in UCT (Equation~\ref{eq:UCT}). Specifically, for each action $a$ grounded by the ASP solver, we increase the values $V(ha)$ to the same value of $c$, and $N(ha)$ to a fixed value representing high reward (this is domain-dependent, we empirically set it to $10$ in our domains). This is the same as implying that we have already performed $N(ha)$ simulations in the current belief, and they achieved a good return value (i.e., an high value of $V(ha)$).
It is important to notice that this prior does not impact the optimality of the policy. In fact, the number $N(ha)$ is finite, thus, as in standard POMCP, performing an infinite number of simulations we would converge toward the optimal policy as in standard POMCP.
However, this prior considerably decreases the need for exploration during MCTS, since the most probable best actions are suggested by axioms before MCTS simulation starts. This is a crucial benefit when the tree grows in size (hence, many simulation are needed), as shown in Section~\ref{sec:exp}.

\section{Empirical evaluation}\label{sec:exp}
We now describe the empirical evaluation for our two case studies, rocksample and battery. 

\subsection{Empirical methodology}\label{sec:exp_setting}
Learning executions are generated running POMCP in 1000 random scenarios for each domain, specifically considering: $12 X 12$ grid with 4 rocks for rocksample, randomizing the value and position of rocks and initial position of the agent;
35-step path for battery, randomizing locations of stations with a limited range of mutual distances ($\in [1, 4]$).
We set $2^{15}$ particles\footnote{As of standard practice, the number of particles is the same as the number of simulations.} and simulations in POMCP algorithm, in order to generate ``good'' executions as explained in Section \ref{sec:ilasp_methods}. 
ILASP examples are extracted only from execution traces with discounted return greater or equal than the average over all traces.
This criterion is domain-independent to select only meaningful examples to learn from.

We evaluate performance of POMCP with and without the bias of ASP axioms in two scenarios: i) using fewer particles with respect to the training set, while keeping problem dimension fixed (50 random settings for each number of particles are generated); ii) increasing the problem dimension with fixed number of particles to $2^{15}$.
Specifically, in the second case we increase the grid size $N$ in rocksample and path length in battery, running 50 random simulations for each parameter value. In this way, we show the scalability of rules to different instances of the domains, and their utility in practical applications, where the planning horizon is critical for POMCP exploration. 
ASP solving is performed with state-of-the-art Clingo \cite{gebser2016theory}.
All experiments are run on a notebook with Intel Core i7-6700HQ and 16GB of RAM.
  
\subsection{Battery}
Results for battery domain for each step of our methodology (Section \ref{sec:methods}) follow.

\subsubsection{ASP representation}
We define the following features for the battery domain:
\begin{itemize}
    \item \stt{guess(L,V)}, representing the probability \stt{V}$\in \{0, 10, ..., 100\}$\% that the level of the battery is at least \stt{L}$\in \{0, 1, ..., 10\}$.
    \item \stt{dist\_next(D)}, representing the distance \stt{D} (integer) from the next station, if any, or the goal.
    \item \stt{X $\geq \Bar{x}$} and \stt{X $\leq \Bar{x}$}, where $\stt{X}$ is either \stt{L,D,V}  and $\Bar{x}$ is a possible value for \stt{X}.
    \item \stt{at\_station}, meaning that the agent is on a station location.
\end{itemize}
Atoms for actions are \stt{recharge, check, advance}.

\subsubsection{ILASP results}
ILASP learns the following causal rules from a set of $\approx 29600$\footnote{The number of examples depends on the number of steps in executions.} examples (i.e., CDPIs as in Section \ref{sec:ilasp_theory}):
\begin{align*}
    &\stt{check :- V} \geq \stt{20; L} \leq \stt{4; guess(L, V).}\\
    &\stt{advance :- dist\_next(D); D} \leq \stt{4.}\\
    &\stt{recharge :- L} \leq \stt{7; V} \geq \stt{30; D} \geq \stt{2; dist\_next(D);}\\ 
    &\ \ \ \ \ \ \ \ \ \ \ \ \ \ \ \ \ \ \ \ \ \ \ \ \ \ \stt{guess(L, V); at\_station.}
\end{align*}
Checking is convenient if the level of battery is considerably low ($\leq 4$) with non-negligible probability ($\geq 20\%$). Recharging should be performed if the level of battery is $\leq 7$ with good probability ($\geq 30\%$), and the next station is not sufficiently close (\stt{D}$\geq 2$). Finally, advancing is always suggested, since the next station (or the goal) is always closer than 4 steps in our problem definition.
The learning time is $\approx$\SI{45}{s} for the longest \stt{recharge} rule (requiring more time), with $S_M$ including $\approx 4700$ axioms\footnote{The size of $S_M$ depends on the number of features, variables and their possible values.}.

\subsubsection{POMCP integration results}
\begin{figure*}
    \centering
    \begin{subfigure}{0.36\textwidth}
    \centering
    \includegraphics[width=\linewidth]{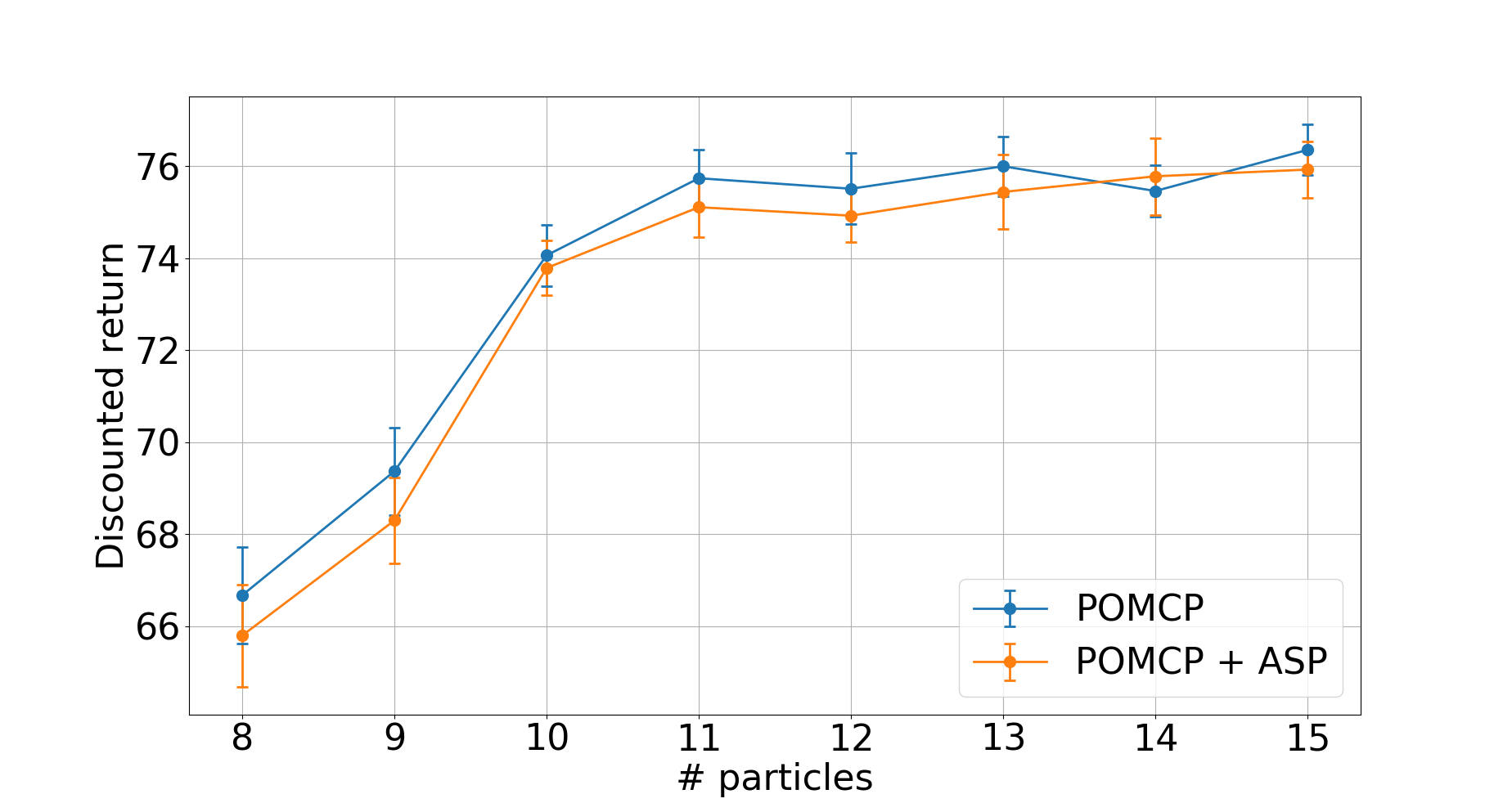}
    \caption{\label{fig:battery_particles}}
    \end{subfigure}
    \begin{subfigure}{0.36\textwidth}
    \centering
    \includegraphics[width=\linewidth]{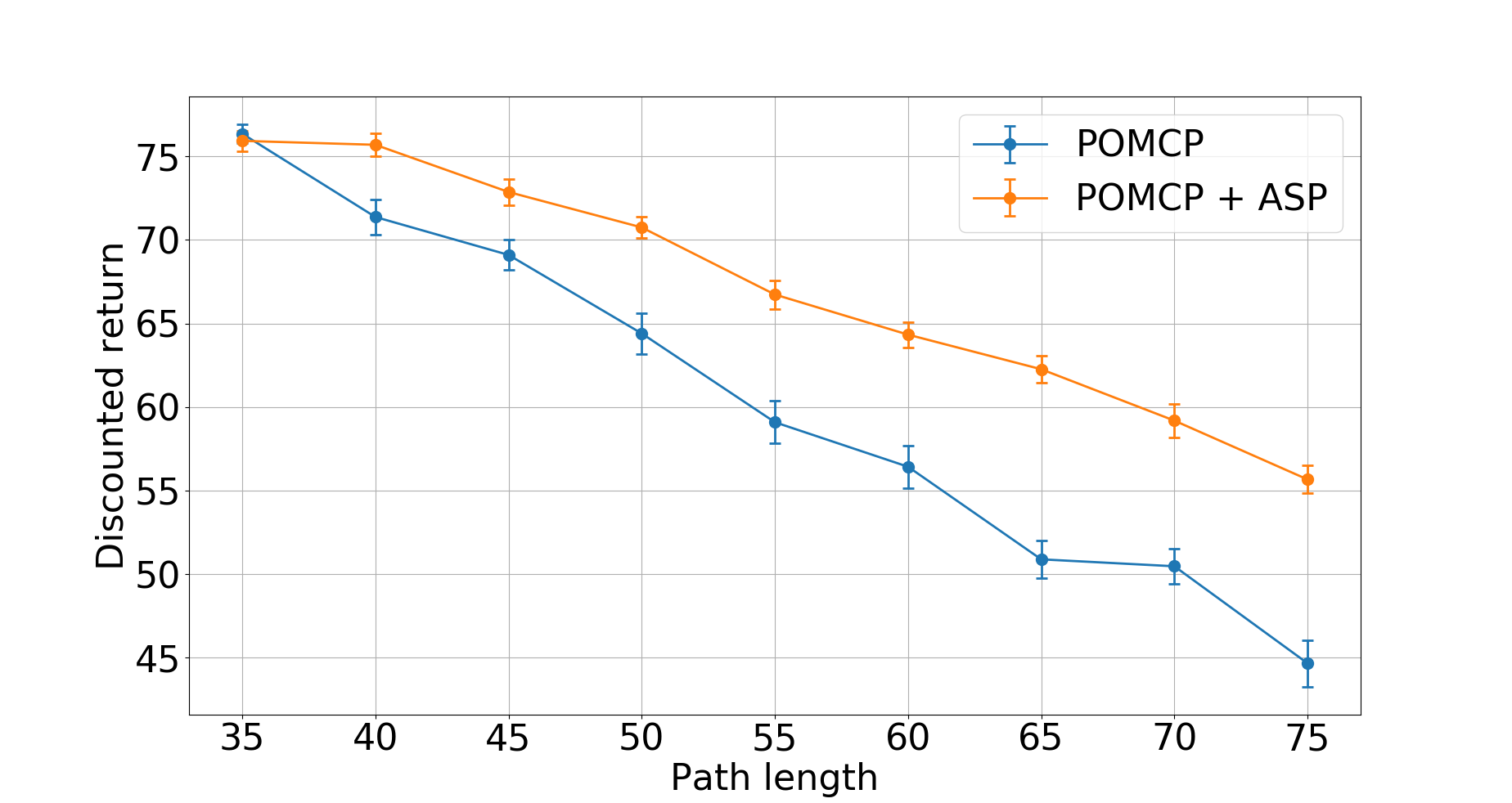}
    \caption{\label{fig:battery_length}}
    \end{subfigure}
    \caption{Battery domain (mean $\pm$ std deviation). a. Tests with reduced number of POMCP particles; b. with increasing path length.}
    \label{fig:results_battery}
\end{figure*}
In Figure \ref{fig:battery_particles}, we show the discounted return achieved by POMCP (with and without rule bias) when the number of particles is varied between $2^{8}$ and $2^{15}$. Introducing learned ASP axioms does not significantly affect the performance of the solver. This is probably due to the limited number of actions involved in the task (only 3, i.e., \stt{advance}, \stt{check} and \stt{recharge}), thus few branches must be explored at each simulation step. 

Figure \ref{fig:battery_length} shows the discounted return achieved by POMCP when the length of the path increases, while the number of particles is constantly $2^{15}$ as in the training setup. In this case, the solver performs significantly better when learned rules are employed (the discounted return increases up to 20\% when the path length is 75, i.e., almost double than the training scenario). In fact, as the path length increases, the risk of battery discharge is higher, since failing to recharge is more probable. Moreover, the planning horizon, hence simulation depth, also increases, so guidance of ASP axioms is crucial.


\subsection{Rocksample}
Results for rocksample domain for each step of our methodology (Section \ref{sec:methods}) follow.

\subsubsection{ASP representation}
We define the following set of features:
\begin{itemize}
    \item \stt{guess(R,V)}, representing the probability \stt{V} that rock \stt{R} is ``good''. Since ASP can only deal with integer variables, \stt{V} is discretized in the range $\{0, 10, ..., 100\}$ with the following meaning: \stt{guess(1, 10)} means that rock \stt{1} is valuable with probability $P(1) \in [10, 20[\ \%$. We choose this discretization step for probabilities empirically; finer discretization is possible, but may affect ASP computational time in large domains.
    \item \stt{dist(R,D)}, representing the 1-norm distance \stt{D} between the agent and rock \stt{R}.
    \item \stt{min\_dist(R)}, meaning that rock \stt{R} is the closest to the agent.
    \item \stt{delta\_x(R,D)} (respectively, \stt{delta\_y(R,D)}), meaning $x$- (respectively, $y$-)coordinate of rock \stt{R} with respect to the agent is \stt{D}.
    \item \stt{X $\geq \Bar{x}$} and \stt{X $\leq \Bar{x}$}, where \stt{X} is either \stt{V,D} and $\Bar{x}$ is a possible value for \stt{X}.
    \item \stt{sampled(R)}, meaning that a rock \stt{R} has been sampled.
    \item \stt{num\_sampled(N)}, meaning that percentage \stt{N} of rocks has been sampled ($\stt{N}\in \{0, 25, ..., 100\}$.
    \item \stt{target(R)}, meaning that rock \stt{R} is the next to be sampled.
\end{itemize} 
In particular, \stt{min\_dist} and \stt{best\_guess} are qualitative features which are added to possibly simplify ASP rule expressions and improve their comprehensibility.

Action atoms are \stt{east}, \stt{west}, \stt{north}, \stt{south}, \stt{exit}\footnote{\stt{exit} is considered as a separate action for generality. Since the agent exits the grid from the right-hand side in our domain, it is manually mapped to \stt{east} action for the scope of this paper, in POMDP representation.}, \stt{check(R)}, \stt{sample(R)}, where argument \stt{R} represents the target rock to be checked or sampled. 

\subsubsection{ILASP results}
ILASP learns the following axioms from a set of $\approx 8500$ examples:
\begin{subequations}
\begin{align}
    &\stt{east :- target(R), delta\_x(R,D), D} \geq \stt{1.}\\
    &\stt{west :- target(R), delta\_x(R,D), D} \leq \stt{-1.}\\
    &\stt{north :- target(R), delta\_y(R,D), D} \geq \stt{1.}\\
    &\stt{south :- target(R), delta\_y(R,D), D} \leq \stt{-1.}\\
    &\stt{target(R) :- dist(R,D), not sampled(R), D}\leq \stt{1.}\\
    &\nonumber\stt{target(R) :- guess(R,V), not sampled(R),}\\
    &\ \ \ \ \ \ \ \ \ \ \ \ \ \ \ \ \ \ \ \ \ \ \ \ \ \ \ \ \stt{70} \leq \stt{V} \leq \stt{80.}\\
    &\nonumber\stt{check(R) :- target(R), not sampled(R),}\\
    &\ \ \ \ \ \ \ \ \ \ \ \ \ \ \ \ \ \ \ \ \ \ \ \ \ \ \ \ \stt{guess(R,V), V} \leq \stt{50.}\\
    &\nonumber\stt{check(R) :- guess(R,V), not sampled(R),}\\
    &\ \ \ \ \ \ \ \ \ \ \ \ \ \ \ \ \ \ \ \ \ \ \ \ \ \ \ \ \stt{dist(R,D), D} \leq \stt{0, V} \leq \stt{80.}\\
    &\nonumber\stt{sample(R) :- target(R), dist(R,D), D} \leq \stt{0,}\\
    &\nonumber\ \ \ \ \ \ \ \ \ \ \ \ \ \ \ \ \ \ \ \ \ \ \ \ \ \ \ \ \stt{not sampled(R), guess(R,V),}\\
    &\label{eq:sample}\ \ \ \ \ \ \ \ \ \ \ \ \ \ \ \ \ \ \ \ \ \ \ \ \ \ \ \ \stt{V} \geq \stt{90.}\\
    &\nonumber\stt{exit :- guess(R,V), V} \leq \stt{40, not sampled(R),}\\
    &\ \ \ \ \ \ \ \ \ \ \ \ \ \ \ \ \ \ \ \stt{num\_sampled(N), N} \geq \stt{25.}\\
    &\nonumber\stt{exit :- dist(R,D), 5} \leq \stt{D} \leq \stt{8, not sampled(R),}\\
    &\ \ \ \ \ \ \ \ \ \ \ \ \ \ \ \ \ \ \ \stt{num\_sampled(N), N} \geq \stt{25.}
\end{align}
\end{subequations}
Axioms for moving actions are quite simple. Motion depends on the target rock and relative position with respect to it (\stt{delta\_x} and \stt{delta\_y} atoms).
Two axioms are found for \stt{target}: an unsampled rock is chosen as a target if either its distance from the agent is low ($\leq 1$) or it is valuable with high probability (guess \stt{V} between $70\%$ and $80\%$). 
Sampling is then performed on the target rock if the agent is at its location and the probability to be valuable is $\geq 90\%$.
Two axioms were discovered also for \stt{check}, which occurs either when the target rock has low ($\leq 50\%$) probability to be valuable, or to verify the value of a rock when on it.
Finally, the agent decides to exit the grid when at least $25\%$ of rocks have been sampled and either an unsampled rock has low probability to be valuable ($\leq 40\%$) or it is far from the agent (distance between 5 and 8).

ILASP is also able to learn the following set of weak constraints for the \stt{target} atom:
\begin{align*}
    &\stt{:}\sim \stt{target(R), dist(R,D).[D@1, R, D]}\\
    &\stt{:}\sim \stt{target(R), min\_dist(R), guess(R,V).}\\
    &\ \ \ \ \ \ \ \stt{[-V@2,R,V]}
\end{align*}
In other words, when multiple rocks can be targets according to causal rules, the agents ranks them considering primarily the closeness, secondly the probability to be valuable.
The learning time is $\approx$\SI{420}{s} for the longest \stt{target} rules (with weak constraints), with $S_M$ including $\approx 29500$ axioms.

\subsubsection{POMCP integration results}
\begin{figure*}
    \centering
    \begin{subfigure}{0.36\textwidth}
    \centering
    \includegraphics[width=\linewidth]{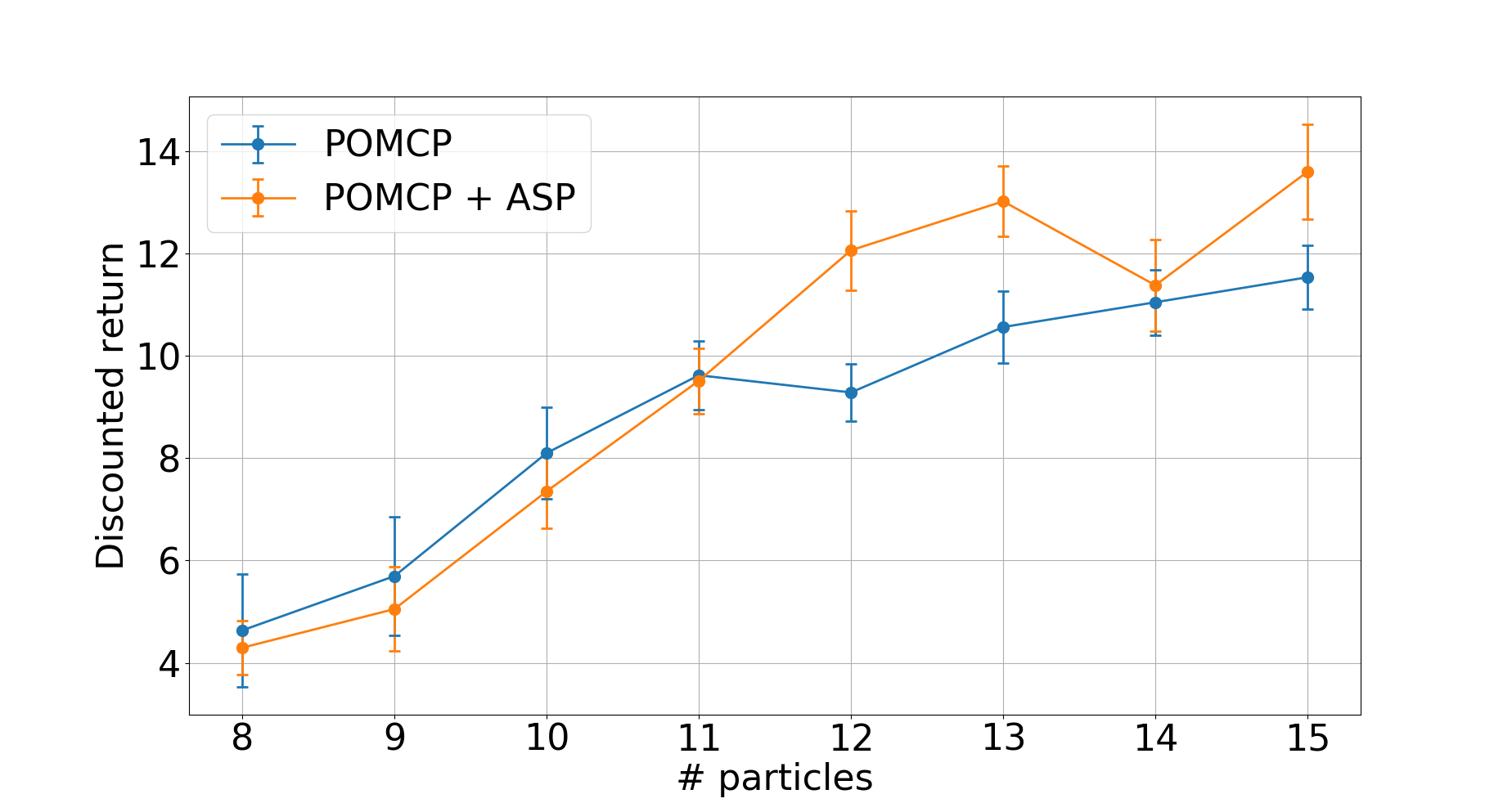}
    \caption{\label{fig:rocksample_particles}}
    \end{subfigure}
    \begin{subfigure}{0.36\textwidth}
    \centering
    \includegraphics[width=\linewidth]{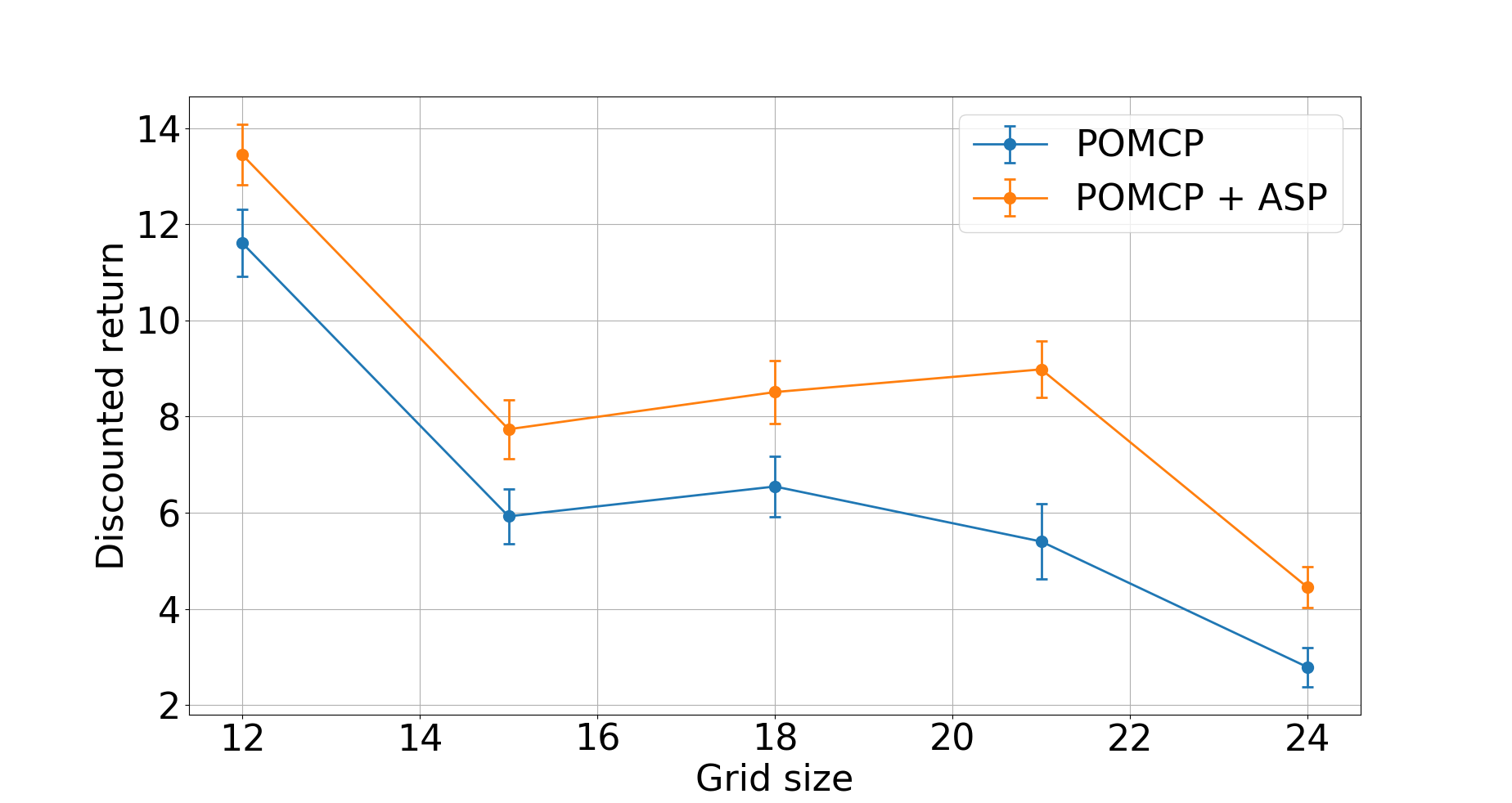}
    \caption{\label{fig:rocksample_size}}
    \end{subfigure}
    \caption{Rocksample domain (mean $\pm$ std deviation). a. Tests with reduced number of POMCP particles; b. with enlarged grid size $N$.}
    \label{fig:results_rocksample}
\end{figure*}
Figure \ref{fig:rocksample_particles} shows the performance of POMCP (with and without rule bias) when the number of particles varies between $2^8$ and $2^{15}$ with grid size and number of rocks as in the training setup ($N=12$ and $M=4$). The discounted returns achieved with and without the rules are very similar to each other when a low number of particles (i.e., $\leq 2^{11}$) is used. When instead the number of particles increases (i.e., $> 2^{11}$), the learned rules improve the performance (up to $15-20\%$ with $2^{15}$ particles)\footnote{The dip at $2^{14}$ particles is due to the lower number of valuable rocks, on average, in the random scenarios.}. In fact, the rules describe more accurately the policy map when a larger amount of particles is used, because the precision of the belief distribution is higher, hence, upper and lower bounds about the distance and probability of rocks to be valuable are more precise (e.g., \stt{D$\leq$0} and \stt{V$\geq$90} in \eqref{eq:sample}). 
Since the number of actions in rocksample is higher than the battery domain,
learned axioms significantly reduce the number of branches to be expanded in the Monte Carlo tree in POMCP exploration. 
This is even more evident in Figure \ref{fig:rocksample_size}, where the grid size is increased from 12 to 24, while the number of rocks and simulations are kept equal to those in the training set, i.e., $M=4$ and $2^{15}$ respectively. In this case, the planning horizon grows, and rules become crucial to achieve better performance.
In fact, POMCP with rules reaches almost double discounted return then standard POMCP, on average, with a $24X 24$ grid.


\subsection{Further considerations}
We now analyze some crucial aspects of our pipeline, specifically the computational impact of answer set solving in POMCP (Section \ref{subsec:disc_comp}) and how the quantity and quality of examples in the training set of executions affect learning and planning outcomes (Sections \ref{subsec:disc_robust}-\ref{subsec:disc_badex}). 
For brevity, we consider only the rocksample scenario, which turned out to be more challenging than battery, since it has more actions.

\subsubsection{Computational impact of ASP in POMCP}\label{subsec:disc_comp}
As explained in Section \ref{sec:pomcp_methods}, ASP solving occurs at every node in MCTS.
In the rocksample scenario with $2^{15}$ particles and 4 rocks on a 12$X$12 grid, Clingo is invoked on average 125000 times per trial, over 50 task instances. Each ASP solving call takes approximately \SI{0.3}{ms}. Compared to pure POMCP solver in the same scenario, this results in higher computational time per step, i.e., executed action ($5.53 s$ vs. $1.88 s$ on average) and over all task instance ($27 s$ vs. $18 s$ on average). Using Clingo has the advantage to keep the implementation modular, so it is easy to modify task axioms in a separate ASP file to be evaluated at runtime. However, it is more efficient to implement ASP axioms directly as conditional statements in POMCP code (an example is provided in the linked repository), resulting in similar computational time as pure POMCP ($20 s$ vs. $18 s$ on average per task instance, in previous conditions).
Moreover, Figure \ref{fig:rocksample_particles} shows that axioms allow to achieve higher discounted return for $\geq 2^{12}$ particles than pure POMCP with $N=2^{15}$. POMCP+Clingo completes a task instance in $\approx 2.5 s$ on average with $2^{12}$ particles, almost one order of magnitude less than pure POMCP with $2^{15}$ particles. Thus, our pipeline can obtain better performance with lower computational cost. 

\subsubsection{Robustness of learned axioms}\label{subsec:disc_robust}
\begin{figure*}
    \centering
    \begin{subfigure}{0.33\textwidth}
    \centering
    \includegraphics[width=\linewidth]{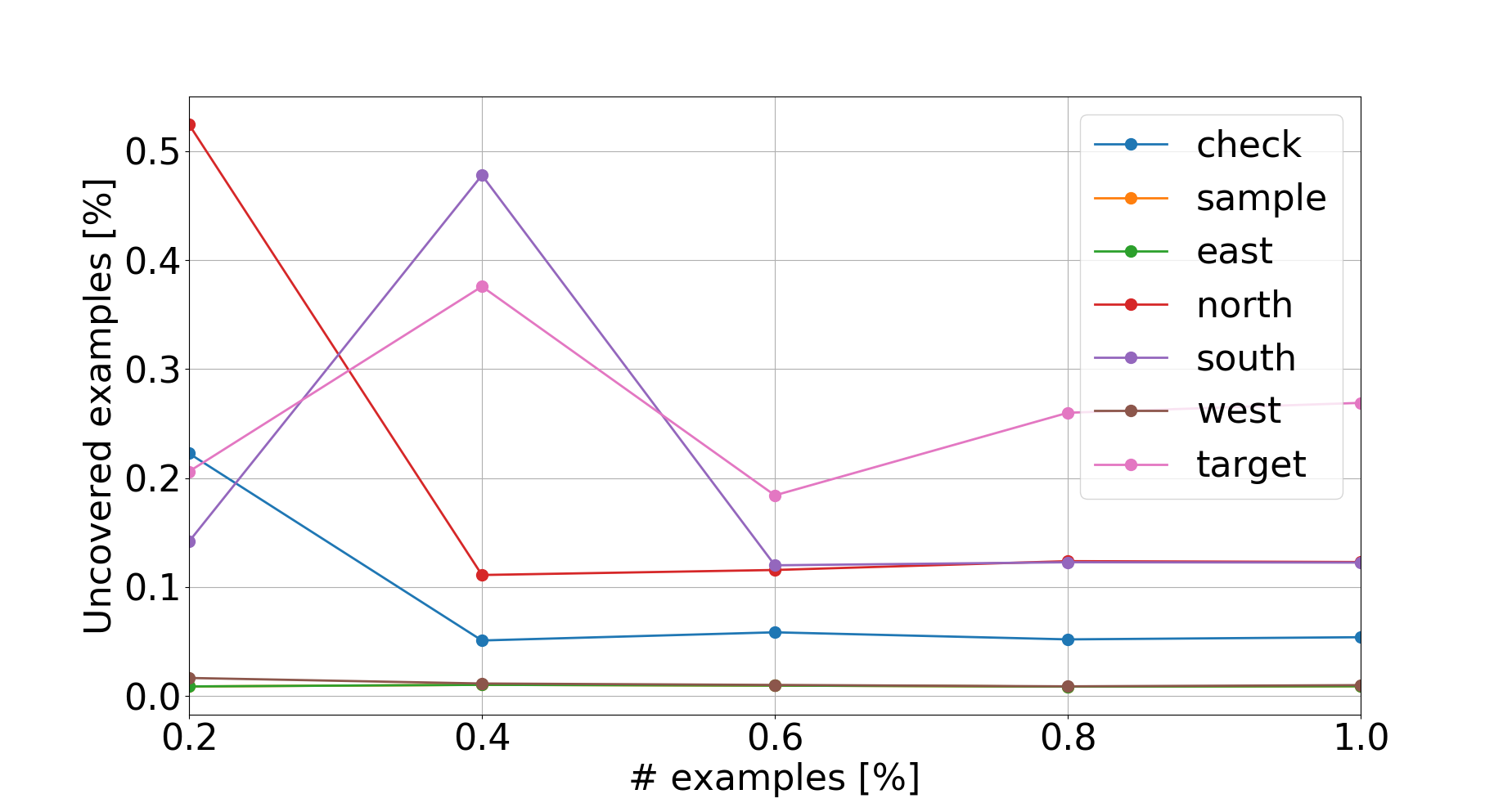}
    \caption{\label{fig:uncovered_ex}}
    \end{subfigure}
    \begin{subfigure}{0.33\textwidth}
    \centering
    \includegraphics[width=\linewidth]{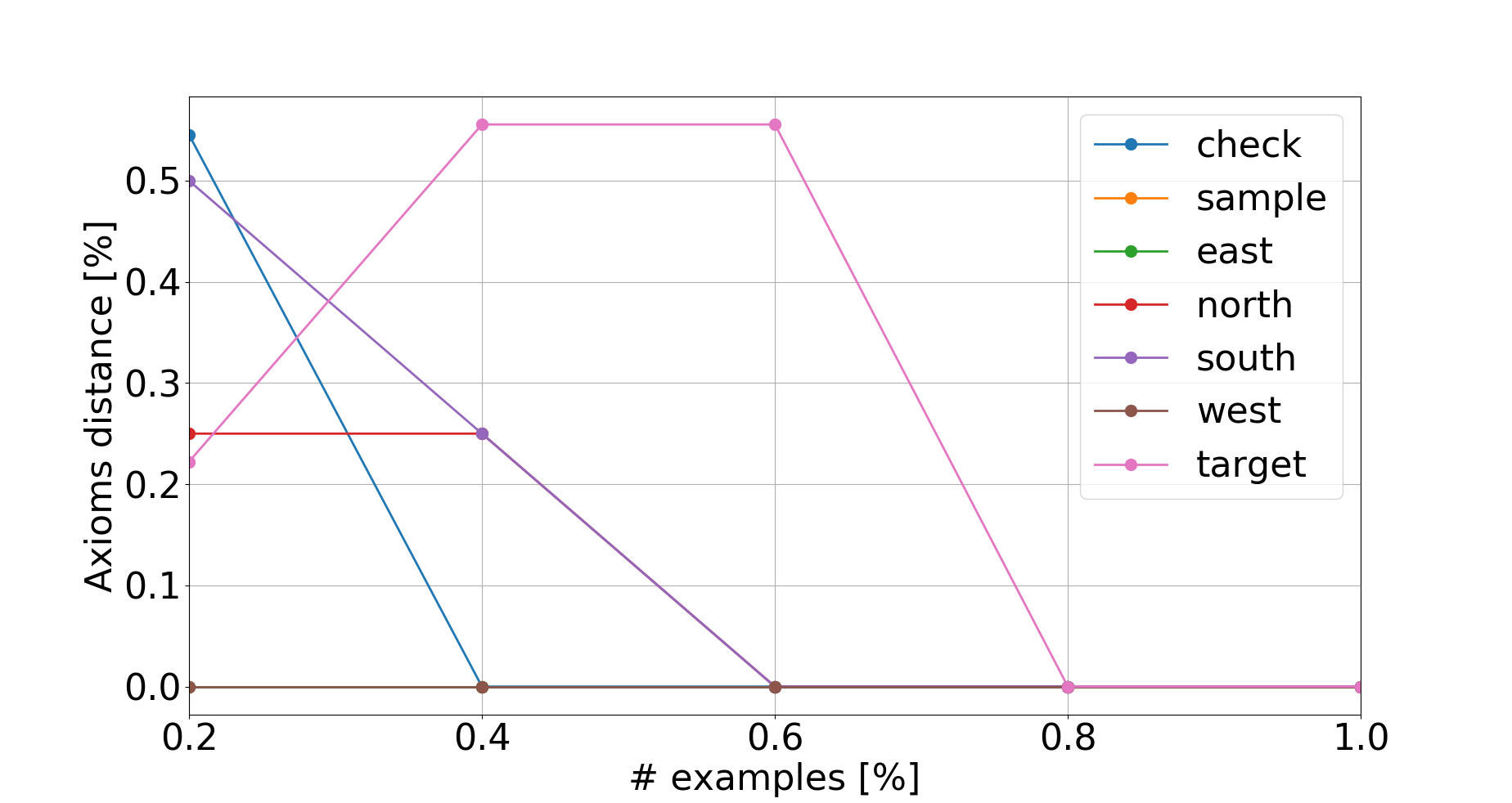}
    \caption{\label{fig:axiom_distance}}
    \end{subfigure}
    \begin{subfigure}{0.33\textwidth}
    \includegraphics[width=\linewidth]{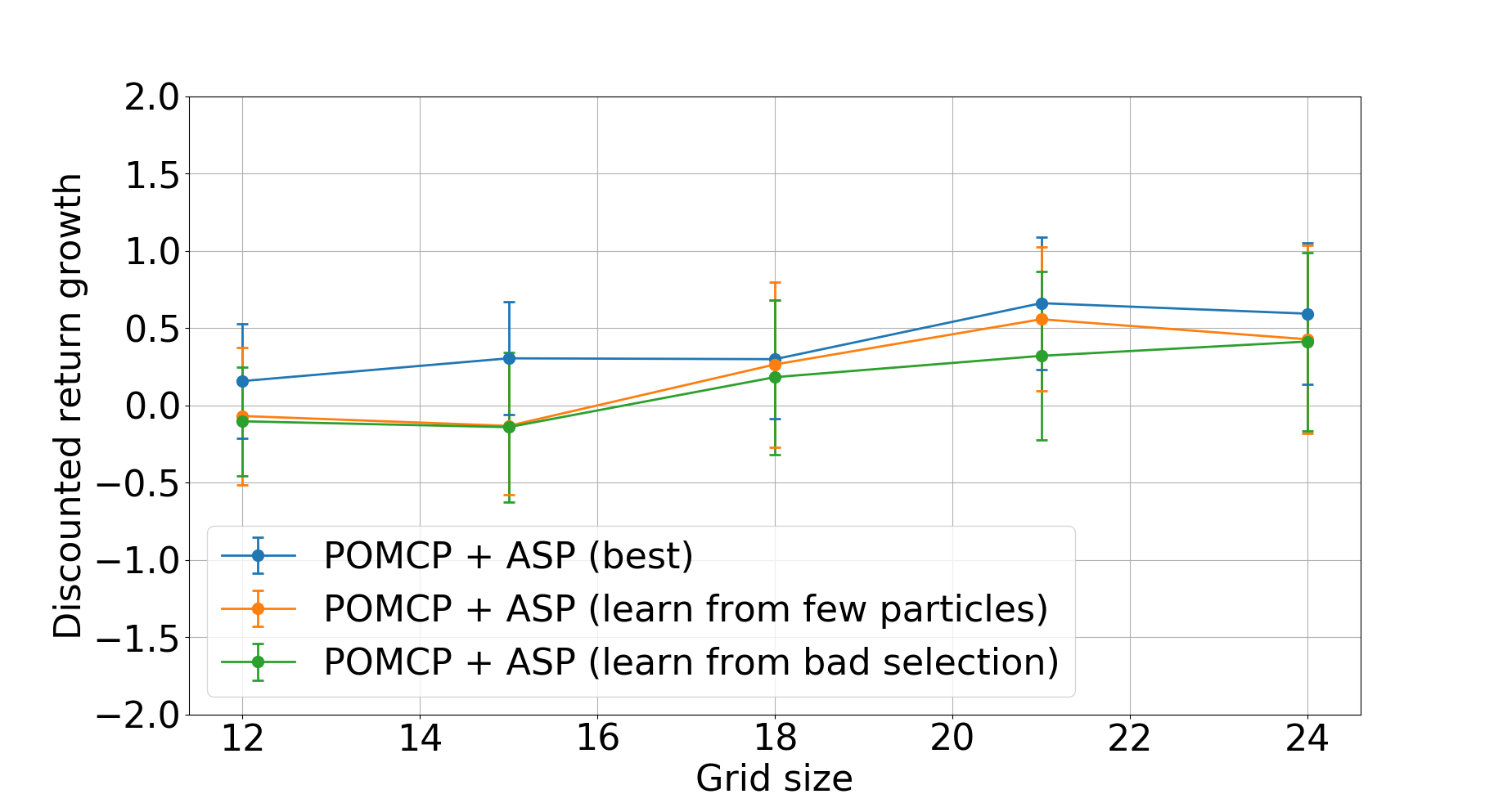}
    \caption{\label{fig:quality_examples}}
    \end{subfigure}
    \caption{a. Non-covered examples (\%) for different number (\#) of examples used for learning; b. Relative distance between rules learned with different \# of examples, w.r.t. rules learned from the full training set; c. Growth rate of discounted return (mean $\pm$ std deviation) for rocksample with larger grid size, when axioms are learned from best examples ($2^{15}$ particles and discounted return > average) or sub-optimal examples (either discounted return < average or generated from POMCP with $2^{11}$ particles).}
    \label{fig:robustness_learning}
\end{figure*}
We now evaluate how the number of examples in the training set for ILASP affects learned axioms.
Figures \ref{fig:uncovered_ex}-\ref{fig:axiom_distance} show the results about rule learning depending on the number of  examples (i.e., CDPIs) used. In the $x$-axis we have the percentage of examples with respect to the original training set used in previous experiments (i.e., $\approx 8500$ in rocksample). In the $y$-axis of Figure \ref{fig:uncovered_ex} we show the \emph{percentage of non-covered examples}, while in the $y$-axis of Figure \ref{fig:axiom_distance} we show the \emph{distance between learned axioms} from a given amount of examples and from the full training set (normalized by the number of atoms in the latter).
Given 2 rules $R_1,R_2$, each one made of a set of atoms $\{\stt{a}_i\}, i\in\{1,2\}$, we define their distance as $R_1 - R_2 = |\{\stt{a}_1\} \cup \{\stt{a}_2\}| - |\{\stt{a}_1\} \cap \{\stt{a}_2\}|$, with $|\cdot|$ the cardinality of a set.
For instance, \stt{sample(R) :- dist(R,V), V}$\leq$\stt{2} has a distance 5 from \eqref{eq:sample}, due to the missing \stt{not sampled(R)}, \stt{guess(R, V),V} $\geq$\stt{90} and \stt{target(R)}, and the different upper bound on distance \stt{D}. 
We observe that using $\geq80\%$ of the full training dataset, the percentage of non-covered examples stabilizes (Figure \ref{fig:uncovered_ex}) and the distance becomes null for all actions (Figure \ref{fig:axiom_distance}), thus learned rules successfully converge to stable hypotheses. 
Overall, rules learned from the full dataset cover more $>73\%$ of examples.

\subsubsection{Influence of quality of examples on POMCP performance}\label{subsec:disc_badex}
We now evaluate the influence of the quality of examples (i.e., of the policy generating them) on the discounted return achieved by POMCP with ASP axioms. 
We want to show what happens when an optimal policy cannot be computed by POMCP for training set generation due to the complexity of the task, so only a sub-optimal one is available.
For testing, we first learn rules for rocksample from POMCP optimal executions generated with $2^{15}$ particles, but selecting only traces with discounted return \emph{smaller} than the average on all traces. This emulates the case in which a suitable criterion for example selection cannot be easily defined, e.g., in complex domains.
Then, in another test we learn rules from traces generated with $2^{11}$ particles, i.e., where axioms improve performance of pure POMCP, and select traces with discounted return > average. This is also useful for fast generation of examples in complex domains.
Figure \ref{fig:quality_examples} shows the results for the two tests (considering 50 random settings for each value of grid size), in terms of the rate of \emph{growth of discounted return}, i.e., the relative difference between discounted return achieved with and without rules, normalized by the discounted return achieved without rules. 
In our charts, POMCP is run with $2^{15}$ particles when combined with rules for validation.
In general, rules learned from sub-optimal examples do not degrade performance with respect to pure POMCP (only a slight average decrease $< 14\%$ is observed for grid size $N < 18$), thanks to the soft guidance approach. Interestingly, sub-optimal ASP rules learned in both tests still are beneficial to POMCP as the planning horizon increases ($N\geq18$). In particular, at $N=24$ discounted return increases of approximately $50\%$.

\section{Conclusion}\label{sec:conclusion}
We presented a novel methodology for soft POMCP policy guidance, based on ILP to learn ASP rules describing policy specifications directly from POMDP execution traces. Our rules do not need to be hand-crafted by experts, but we only require definition of relevant high-level domain-specific features, in order to describe belief information in ASP formalism and find matches with POMDP actions. We showed that learned ASP axioms significantly support POMCP exploration in the benchmark domains of rocksample and battery, advising good branches to be explored in POMCP simulation without significantly affecting the computational burden. In particular, rules are more beneficial when the number of possible actions increases (rocksample) and the planning horizon is extended. Furthermore, POMCP property of asymptotic optimality is preserved when axioms are learned from traces generated with low-quality policies, and axioms converge to a fixed point as the number of examples increases.

In the future, we plan to investigate i) how to integrate more advanced (e.g., temporal) logic specifications to address more complex tasks; ii) how to account for non-covered examples in the training set and possibly refine axioms during execution; iii) the influence of bad defintion of environmental features by users.


\begin{acks}
This project has received funding from the Italian Ministry for University and Research, under the PON “Ricerca e Innovazione” 2014-2020" (grant agreement No. 40-G-14702-1).
\balance
\end{acks}



\bibliographystyle{ACM-Reference-Format} 
\bibliography{sample}


\end{document}